\documentclass{article}

\PassOptionsToPackage{numbers, compress}{natbib}

\usepackage[preprint]{neurips_2023}

\usepackage[utf8]{inputenc} 
\usepackage[T1]{fontenc}    
\usepackage{url}            
\usepackage{booktabs}       
\usepackage{amsfonts}       
\usepackage{nicefrac}       
\usepackage{microtype}      
\usepackage{xcolor}         
\usepackage{colortbl}
\usepackage{subcaption}
\usepackage{enumitem}
\usepackage{graphicx}
\usepackage{multirow}
\usepackage{amsmath}
\usepackage{makecell} 
\usepackage{amsfonts,amssymb}
\usepackage{pifont}       
\usepackage{bbding}       
\usepackage{fontawesome}  
\usepackage{wrapfig}
\usepackage{tcolorbox}

\newcommand{\cmark}{\ding{52}}
\newcommand{\xmark}{\ding{55}}

\definecolor{my_green}{RGB}{51,102,0}
\definecolor{my_yellow}{RGB}{255,165,0}
\definecolor{my_red}{RGB}{204, 0, 0}

\newcommand{\colorcmark}{\textcolor{my_green}{\ding{52}}}
\newcommand{\colorxmark}{\textcolor{my_red}{\ding{55}}}

\definecolor{hotpink}{RGB}{59, 115, 227}
\usepackage[pagebackref=true,breaklinks=true,letterpaper=true,colorlinks,bookmarks=false,citecolor=hotpink]{hyperref}

\title{GPT4Tools: Teaching Large Language Model to Use Tools via Self-instruction}

\author{
Rui Yang$^1$\footnotemark[1]\, \footnotemark[3],\quad 
Lin Song$^2$\footnotemark[1]\, \footnotemark[2],\quad
Yanwei Li$^3$,\quad
Sijie Zhao$^2$,\quad
Yixiao Ge$^2$,\quad
Xiu Li$^1$,\quad
Ying Shan$^2$\\
$^1$Tsinghua Shenzhen International Graduate School, Tsinghua University\\
$^2$Tencent AI Lab\quad $^3$Chinese University of Hong Kong\\
{\tt\small rayyang0116@gmail.com} \quad
{\tt\small ronnysong@tencent.com}
}

\begin{document}

\renewcommand{\thefootnote}{\fnsymbol{footnote}}
\footnotetext[1]{Equal contribution. $\ddagger$ Work done during an internship at Tencent.}
\footnotetext[2]{Corresponding author.}
\renewcommand{\thefootnote}{\arabic{footnote}}

\maketitle

\emph{The essential difference between humans and animals is that humans are capable of making and using tools.}

\rightline{---Friedrich Engels}

\begin{abstract}
This paper aims to efficiently enable Large Language Models (LLMs) to use multimodal tools.
Advanced proprietary LLMs, such as ChatGPT and GPT-4, have shown great potential for tool usage through sophisticated prompt engineering.
Nevertheless, these models typically rely on prohibitive computational costs and publicly inaccessible data.
To address these challenges, we propose the GPT4Tools based on self-instruct to enable open-source LLMs, such as LLaMA and OPT, to use tools.
It generates an instruction-following dataset by prompting an advanced teacher with various multi-modal contexts.
By using the Low-Rank Adaptation (LoRA) optimization, our approach facilitates the open-source LLMs to solve a range of visual problems, including visual comprehension and image generation.
Moreover, we provide a benchmark to evaluate the ability of LLMs to use tools, which is performed in both zero-shot and fine-tuning ways.
Extensive experiments demonstrate the effectiveness of our method on various language models, which not only significantly improves the accuracy of invoking seen tools, but also enables the zero-shot capacity for unseen tools.
The code and demo are available at \href{https://github.com/StevenGrove/GPT4Tools}{https://github.com/StevenGrove/GPT4Tools}.
\end{abstract}

\section{Introduction}

Recent advances in large language models (LLMs), such as GPT-3~\cite{GPT3}, InstructGPT~\cite{InstructGPT}, and ChatGPT~\cite{ChatGPT}, have demonstrated substantial potential in the area of zero-shot learning and logical reasoning.
These models are typically trained on a large volume of text-only data, primarily sourced from the internet.
However, as promising as they may seem, these advanced proprietary LLM~\cite{ChatGPT, GPT4} have significant limitations. 
One of the major hindrances is the high computational cost associated with these models, which may not be affordable or accessible to many scenarios.
Additionally, these models typically depend on specialized data, such as source code and conversation history, which are not easily available to the public.

Instead of solely focusing on language processing, many recent researches~\cite{VisualChatGPT, MMREACT} attempt to bridge the gap between language models and multi-modal tools.
The intelligent agents like Visual ChatGPT~\cite{VisualChatGPT} and MMREACT~\cite{MMREACT} have made efforts to meet this goal by sophisticated prompt engineering.
These agents utilize a pre-defined template to create instructions that can be executed by vision-language foundation models.
Although these approaches have led to impressive results, the primary process of instruction decomposition is heavily based on GPT-3.5~\cite{ChatGPT}, which is not publicly available, thus limiting further advancements.
In addition, equipping these agents with the capability to use tools requires a large amount of data~\cite{ChatGPT}. 
This brings up an open question: \textit{how to efficiently enable a primitive language models to use multi-modal tools?}

To achieve it, different from previous studies~\cite{komeili2021internet, WikiText, lamda, lazaridou2022internet, toolformer}, we explore a new perceptive as illustrated in Table~\ref{tab:statistic}.
We propose a simple yet effective method, called GPT4Tools, designed to empower open-source LLMs with the ability to use tools via self-instruct from advanced LLMs.
To achieve this, we construct an instruction dataset by prompting advanced teachers (for example, ChatGPT~\cite{ChatGPT}) conditional on visual content and tool descriptions, which leads to the generation of tool-related instructions.
Unlike Toolformer~\cite{toolformer}, our method can utilize visual content description to significantly improve data diversity.
Furthermore, with the generated instruction-following dataset, we incorporate Low-Rank Adaptation (LoRA) to fine-tune the primitive language models, such as Vicuna~\cite{vicuna} and OPT~\cite{OPT}.
Besides intrinsic language abilities, by using GPT4Tools, language models can also have the capability to use tools to solve a variety of visual problems.
The tasks include visual comprehension and image generation, such as object grounding and segmentation, generating and instructing images, and visual question answering (VQA).
With the proposed GPT4Tools, the LLMs not only significantly improves the accuracy of invoking seen tools, but also enables the zero-shot capacity for unseen tools in a zero-shot manner.

We propose an evaluation metric to assess the effectiveness of LLMs in utilizing tools across diverse tasks.
With this metric, two human-curated validation sets are constructed to evaluate the LLMs in zero-shot and fine-tuning ways, providing a comprehensive measure of the ability to use tools. 
To demonstrate the effectiveness of GPT4Tools, we conduct extensive experiments on various language models.
The results show the efficacy in teaching LLMs when and how to use tools.
Specifically, with the GPT4Tools, the fine-tuned Vicuna-13B achieves 9.3\% absolute gains in successful rate over GPT-3.5~\cite{ChatGPT}, which acquires tool priors in context.
Moreover, the fine-tuned Vicuna-13B shows the strong capacity to invoke unseen tools, which is comparable to GPT-3.5 in successful rate.

Our GPT4Tools stands distinct from previous and concurrent studies~\cite{VisualChatGPT, MMREACT,komeili2021internet, WikiText, lamda, lazaridou2022internet, toolformer} in three ways.
First, our method enables primitive open-source language models to use tools, eliminating the dependence on advanced proprietary LLMs like ChatGPT.
Second, we design a new approach based on multi-modal contexts for self-instruction and augmentation, which significantly promote the multi-modal tool usage and can be deployed in different approaches.
Third, we propose a new benchmark to assess the effectiveness of using tools, and our method shows remarkable improvements.

\begin{table}[]
\caption{Comparison of related works. `LM' is the language models. `Mechanism' denotes how the language model learns to invoke tools. `Unseen' indicates the zero-shot capability on unseen tools.}
\renewcommand\arraystretch{1.2} 
\resizebox{\columnwidth}{!}{
\begin{tabular}{lccccc}
\Xhline{1.0pt}
\textbf{Method}        & \textbf{LM}             & \textbf{Mechanism}              & \textbf{Teacher} & \textbf{Multi-Modal} & \textbf{Unseen} \\ \hline
\citet{lazaridou2022internet}             & Gopher-280B~\cite{Gopher}    & prompt             & \colorxmark                                                               & \colorxmark   & \colorxmark                                                             \\
ToolFormer~\cite{toolformer}    & GPT–J (6B)~\cite{gpt-j}     & self-instruct   & \colorxmark                                                               & \colorxmark   & \colorxmark                                                             \\
Visual ChatGPT~\cite{VisualChatGPT} & GPT-3.5 (175B)~\cite{ChatGPT} & prompt             & \colorxmark                                                               & \colorcmark     & \colorcmark                                                             \\
MMREACT~\cite{MMREACT} & GPT-3.5 (175B)~\cite{ChatGPT} & prompt             & \colorxmark                                                               & \colorcmark     & \colorcmark                                                             \\
GPT4Tools (\textbf{ours})     & Vicuna-13B~\cite{vicuna}     & self-instruct & \colorcmark                                                               & \colorcmark     & \colorcmark                                                             \\ \Xhline{1.0pt}
\end{tabular}}
\label{tab:statistic}
\end{table}

\section{Related Work}
\textbf{Vision and Language Model.}
In the quest to achieve multimodal models capable of addressing both language and vision tasks, several studies~\cite{Caption_LLM_1, GPT3_caption, Uni-Perceiver, Unified-io, Pixel2Pixel, Gato} have explored methods to enable language models to comprehend visual input. These include techniques such as transforming images into discrete textual representations~\cite{Caption_LLM_1, GPT3_caption} or projecting continuous image features into the textual feature space~\cite{BLIP, BLIP2, Flamingo, driess2023palm}. Concurrently, other research has been dedicated to the development of generalist models~\cite{Uni-Perceiver, Unified-io, Gato, Pixel2Pixel}, which permit a model to simultaneously input images and text, eliminating the necessity for a projection process. For instance, OFA~\cite{OFA} devised a unified sequence-to-sequence decoding architecture applicable to language and object detection tasks. Similarly, Pixel2Pixel~\cite{Pixel2Pixel} converted the outcome of visual comprehension tasks into a series of discrete tokens akin to language tasks. Gato~\cite{Gato} brought together a range of vision and control tasks into a sequential prediction issue, while UViM~\cite{UviM} and Unified-IO~\cite{Unified-io} advocated for the learned discrete codes as a means to unify an array of vision tasks.
By contrast, we in this paper equip the language model with a diverse array of specialized multi-modal tools to process distinct vision tasks. This approach not only promotes the scalability of the model for various tasks, but also avoids the issue of forgetfulness stemming from repeated fine-tuning.

\textbf{Instruction Tuning.} 
Recent studies~\cite{meta_tuning, InstructGPT, Zero_Shot_Learners, FLANT5, PromptSource, OPT-IML} have turned out that pre-trained language models could follow natural language instructions and complete various real-world tasks if they are tuned on specific instruction-following data. Notably, InstructGPT~\cite{InstructGPT}, FLAN-T5~\cite{FLANT5}, OPT-IML~\cite{OPT-IML} demonstrated remarkable performance on specific tasks after being fine-tuned with instruction data. In order to release the cost of human-written instructions, Self-Instruction~\cite{self_instruction} found that the instruction-following capabilities of language models can be enhanced by turning on their own generated instruction data. More importantly, this approach inspired a feasible means to improve the zero- and few-shot abilities of language models, i.e., distilling off-the-shelf language models using instructional data from strong ChatGPT~\cite{ChatGPT} or GPT-4~\cite{GPT4}. As a result, many recent works~\cite{vicuna, alpaca, minigpt4, llava} tried to construct excellent language models for various applications based on the LLaMA~\cite{llama}. For instance, Stanford-Alpaca has employed 52K instructions generated by GPT-3.5~\cite{ChatGPT} to construct an exceptional dialogue model. LLaVa~\cite{llava} has adopted GPT-3.5~\cite{ChatGPT} and GPT-4~\cite{GPT4} to incorporate instruction-following data related to visual content. In this paper, we use GPT-3.5~\cite{ChatGPT} to construct tool-related instruction datasets, thereby allowing other language models to acquire tool usage capabilities.

\begin{figure}[t]
  \centering
  \includegraphics[width=\linewidth]{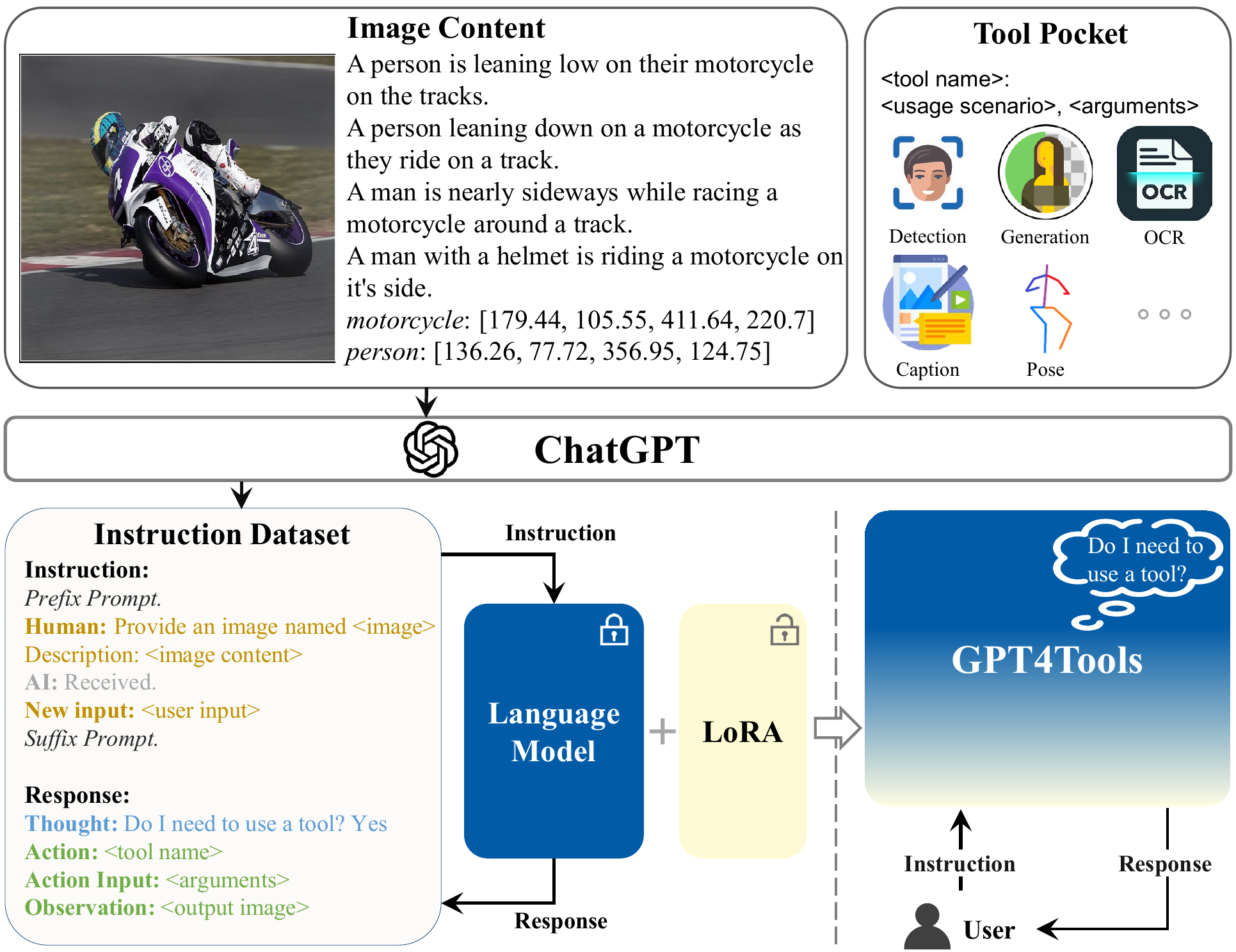}
   \caption{
   Diagram of the GPT4Tools. We prompt the ChatGPT with image content and definition of tools in order to obtain a tool-related instruction dataset. Subsequently, we employ LoRA~\cite{LoRA} to train an open source LLM on the collected instruction dataset, thus adapting the LLM to use tools.
   }
   \label{fig:model_flow}   
\end{figure}

\textbf{Tool Usage.} 
In the Natural Language Processing (NLP) community, several arts~\cite{komeili2021internet, WikiText, lamda, lazaridou2022internet, toolformer} sought to endow language models with the ability to use tools. For instance, \citet{komeili2021internet} proposed to generate conversation responses conditioned on the results of the search engine. LaMDA~\cite{lamda} created a set of tools (comprising an information retrieval system, a calculator, and a translator) to avoid plausible outputs. ~\citet{lazaridou2022internet} utilized few-shot prompting on Gopher-280B~\cite{Gopher} to enable the search engine to ground its output in factual and current information. Similarly, Visual ChatGPT~\cite{VisualChatGPT} and MMREACT~\cite{MMREACT} prompted ChatGPT to invoke visual foundation models. In addition, ToolFormer~\cite{toolformer} used self-instruction and bootstrapping to teach GPT-J (6B)~\cite{gpt-j} using five tools, which include a question and answer system, a calculator, a search engine, a machine translation system, and a calendar. On the contrary, we focus on using the GPT-3.5 model as a powerful teacher to distill off-the-shelf language models and enable them to access many visual models.

\section{Method}

Large language models (LLMs)~\cite{GPT3, OPT, Gopher} have shown remarkable in-context learning abilities. Among them, ChatGPT~\cite{ChatGPT} and GPT-4~\cite{GPT4} are proven to effectively perform text-annotation tasks~\cite{gilardi2023chatgpt} or instruct other models to follow instructions of specific domains~\cite{peng2023instruction, alpaca, vicuna, llava}. Inspired by these findings, we propose leveraging ChatGPT as a powerful teacher to enable off-the-shelf language models to acquire tool usage capabilities. Specifically, we utilize ChatGPT to generate tools-related instruction-following data, which is then used to tune the language model. This process enables the language model to access multimodal information by invoking visual models. Furthermore, we propose an evaluation metric to assess the tool-use ability of the given language model. In the following, we present the data generation, instruction tuning, and evaluation metric in turn.

\subsection{Dataset Construction}
\label{sec:data_construction}
\begin{figure}[t]
  \centering
  \includegraphics[width=0.8\linewidth]{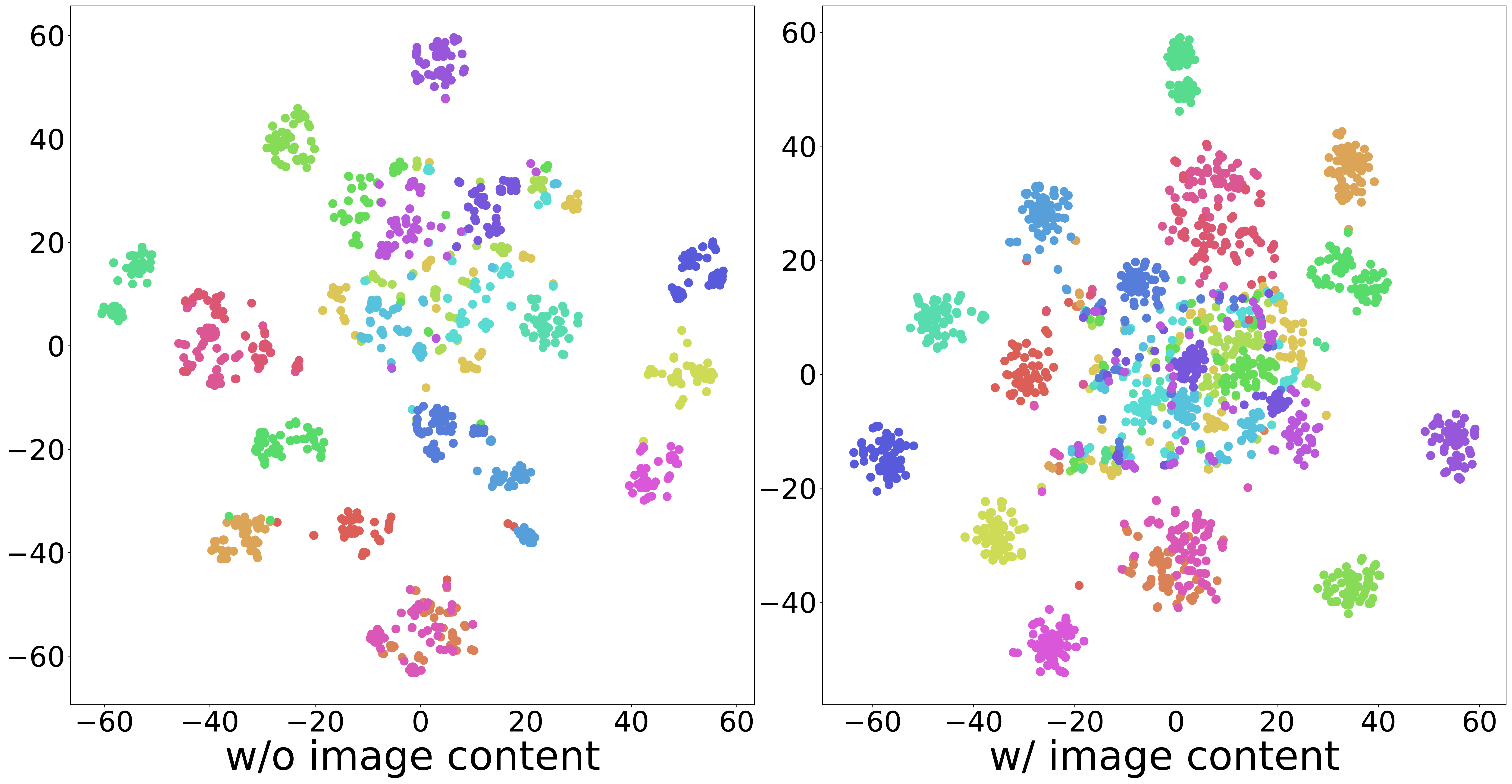}
   \caption{t-SNE\protect\footnotemark visualization for instruction data with and without image content.}
   \label{fig:data_analysis}   
\end{figure}

\textbf{Data Generation.} Figure~\ref{fig:model_flow} illustrates the process of generating tool-related instruction dataset. Given an image, we construct the image content $X_C$ according to the captions and bounding boxes, which is a straightforward means of establishing connections between an image and a language model~\cite{Caption_LLM_1, GPT3_caption}. Conditioned upon the $X_C$, we provide the ChatGPT~\cite{ChatGPT} ($\mathrm{M_T}$) with a tool-related prompt $P_t$ whereby attaining a large number of instruction-following data:
\begin{equation}
    Y \sim \mathrm{M_T}(P_t|X_C).
\end{equation}
The $P_t$ comprises the system message, the definition of tools ($\texttt{<tool name>}:$ $\texttt{<usage scenario>}$, $\texttt{<arguments>}$), and suffix prompt which encourage $M_T$ to generate visual instructions and desired outputs. $Y$, the outcome of $\mathrm{M_T}$, consists of $N$ instruction-output pairs $\{y^1, y^2, ..., y^N\}$, where $y_i$ has the format of "$\texttt{<instruction>},$ $\texttt{<tool name>},$ $\texttt{<arguments>}$", and $N$ is the number of defined tools.
As each input of $\mathrm{M_T}$ is grounded to the image content $X_C$, the generated instructions are inherently connected to the image, thus avoiding arbitrary generation. Moreover, the rich variability of the image results in a higher diversity of instructions when compared to imagined ones. To provide contrast, we also collect instruction follow-up data without image content $X_C$, which is similar to ToolFormer~\cite{toolformer}. As depicted in Figure~\ref{fig:data_analysis}, the high degree of similarity observed among instructions generated without $X_C$ highlights their monotony and lack of diversity. On the contrary, instructions generated with image-conditioned prompts are more informative and diverse, due to changes in the image content.

\footnotetext{The visual instruction and tool arguments are embedded by Sentence-BERT~\cite{ScentenceBERT}. During generation without image content, we set the temperature as $0.9$ to avoid constant outcomes.}

\begin{figure}[t]
  \centering
  \includegraphics[width=\linewidth]{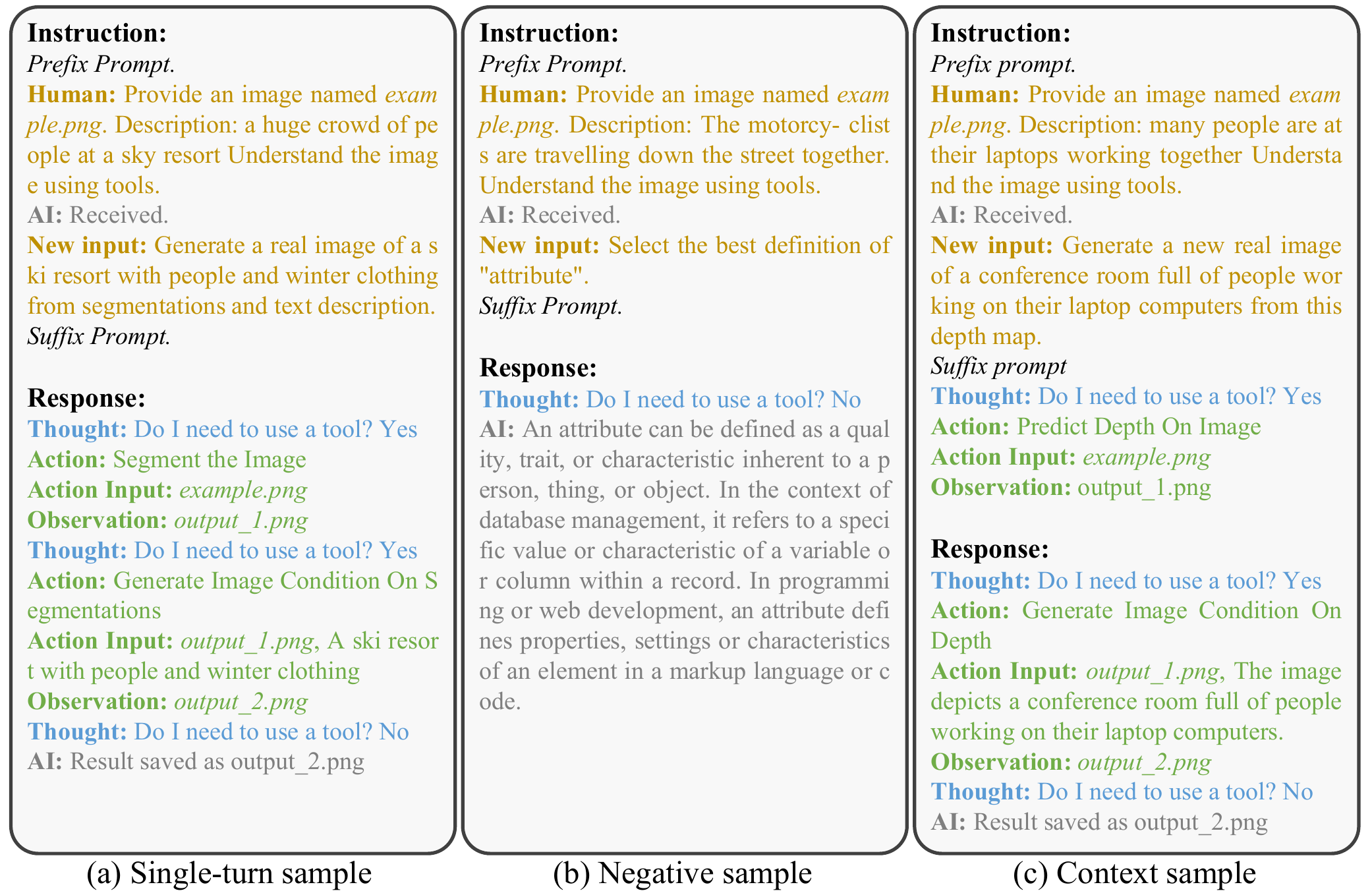}
   \caption{
   Samples of the single-turn instruction, negative instruction, and contextual instruction.  
   }
   \label{fig:samples}   
\end{figure}

\textbf{Data Formation.} 
Upon the collected raw dataset ($70K$ items), we apply a filtering process to remove similar instructions, resulting in $41K$ retained items. Subsequently, we transform the retained data into an instruction-response format utilizing a standardized template as shown in the bottom-left corner of Figure~\ref{fig:model_flow}. This procedure produces a new dataset, denoted as $Y_S^{+}$. The instruction component of $Y_S^{+}$ incorporates a $\textit{prefix prompt}$ that encompasses system messages and tool definitions, $\texttt{<image content>}$ that denotes the image content, $\texttt{<user input>}$ that is replaced with the generated visual instruction, and a $\textit{suffix prompt}$ designed to prompt the language model to reply the user input through given tools. The response component of $Y_S^{+}$ comprises the $\textit{Thought}$ to invoke tools and a chain of actions. Each action involves an $\textit{Action}$ and $\textit{Action Input}$, succeeded by $\texttt{<tool name>}$ and $\texttt{<arguments>}$, respectively. The $\textit{Observation}$ reflects the outcome of the invoked tools.
A sample from $Y_S^{+}$ is presented in the Figure~\ref{fig:samples} (a).

\textbf{Data Augmentation.} Although we have successfully acquired instruction-following data related to the tool usage, this simplistic format lacks complexity and depth in both instructions and responses. To mitigate this issue, we augment the generated data from two perspectives:
\begin{itemize}[leftmargin=*]
    \item \textit{Negative samples.} The generated instructions primarily focus on tool usage, i.e., the decision after the $\textit{Thought}$ is always "Yes". Consequently, there is a potential risk that the fine-tuned model overfits such a decision.  When the user instruction does not connect with the tool usage, the fine-tuned model may erroneously execute irrelevant actions by invoking unnecessary tools. To mitigate this issue, we synthesize negative samples $Y_S^{-}$ by selecting conversation data from the existing dataset~\cite{peng2023instruction} and converting them into the required template, as illustrated in Figure~\ref{fig:samples} (b). By tuning the model with $Y_S^{+} \cup Y_S^{-}$, it can decide when to use tools.
    \item \textit{Context samples.} The generated instructions adopt a standard and fixed single-tune format, which lacks a contextual structure. Thus, we augment the dataset by cutting off the chain of action, as shown in Figure~\ref{fig:samples} (c). Furthermore, we randomly select multiple instructions from $Y_S^{+} \cup Y_S^{-}$ and reformat them into multi-turn conversation data. In this way, we synthesize the contextual instruction-following data $Y_S^{c}$, which enables the tuned model to call tools within the given context.
\end{itemize}

So far, we have constructed the tool-related instructional dataset, including positive samples, negative samples, and context samples: $Y_S=Y_S^{+} \cup Y_S^{-} \cup Y_S^{c}$. 

\subsection{Instruction Tuning}
Based on the $Y_S$, we tune the off-the-self language model using its original auto-regressive training objective. To make the tuning feasible, we leverage LoRA~\cite{LoRA} optimization,, which freezes the language model and optimizes rank decomposition components of the Transformer layers. For a sequence with $L$ tokens, we compute the probability of the target response $X_r$ by:
\begin{equation}
    p(X_r|X_C, X_{inst}) = \prod_{i=1}^{L} p_{\theta}(x_i|X_C, X_{inst}, x_{1:i-1}),
\end{equation}
where $X_{inst}$ denotes the instruction tokens; and $\theta$ is the trainable parameters. In practice, $\textit{prefix prompt}$ and $\textit{suffix prompt}$ are also involved, but we here skip them for better readability.

\subsection{Evaluation Approach}

Numerous benchmarks~\cite{ImageNet,COCO,WikiText} typically utilize human-annotated datasets to evaluate the performance of a model.
For the purpose of measuring the tool-usage capacity of the language model, we construct a evaluation dataset following the same procedures detailed in \S~\ref{sec:data_construction}, and and manually verify the accuracy of each constituent item.
This evaluation dataset is partitioned into two components: the first part (validation set) has the same ingredients as the training set, encompassing 23 tools; the second part (test set) comprises 8 novel tools that are absent from the training set. 
We will use the validation set to validate whether the model can adhere to user commands correctly after tuning with the training set. The test set will be employed to verify whether the model can generalize to new tools after tuning.
Based on the human-annotated evaluation dataset with $N$ instructions, we design a successful rate to measure the model’s performance from three aspects:

\begin{itemize}[leftmargin=*]
\item \textbf{Successful Rate of Thought} (${\mathrm{SR}_t}$) measures whether the predicted decision matches the ground-truth decision. It is calculated as $\mathrm{SR}_{t} = \frac{1}{N} \sum_{i=1}^{N} \mathbb{I}(\tau_i)$,
where $\tau_i$ signifies a singular process. If the thought is correct, $\mathbb{I}(\tau_i)$ is equal to $1$, and $0$ otherwise.
\item \textbf{Successful Rate of Action} ($\mathrm{SR}_{act}$) measures whether the predicted tool name is in agreement with the name of the ground truth tool. It is calculated as $\mathrm{SR}_{act} = \frac{1}{N} \sum_{i=1}^{N} \mathbb{I}(\alpha_i)$, where $\alpha_i$ denotes the matching process for the tool names. In cases where the predicted tool name matches the pre-defined name, $\mathbb{I}(\alpha_i)$ is equal to 1, and 0 otherwise.
\item \textbf{Successful Rate of Arguments} ($\mathrm{SR}_{args}$) evaluates whether the predicted arguments match the ground-truth arguments. 
It can be calculated using the following equation:
\begin{equation}
    \mathrm{SR}_{args} = \frac{1}{N} \sum_{i=1}^{N} \eta_{i},~\mathrm{where}~\; \eta_i = \frac{1}{K} \sum_{j}^{K} \eta_{i, j}.
\end{equation}
Here, $\eta_i$ denotes a sequence of arguments, encompassing both the image path and the input text. For instance, ControlNet~\cite{controlnet} needs the image path saved conditions (e.g. pose map, depth map, or segment map) and the input text described the user command. $K$ represents the quantity of arguments in $\eta_i$. When the argument belongs to the image path, $\eta_{i, j}$ equals $1$ if the predicted and ground-truth image paths share the same suffix, and $0$ otherwise.
When the argument is the input text, $\eta_{i,j}$ is equal to the BLEU score between the predicted and the ground truth text. 

\item \textbf{Successful Rate} ($\mathrm{SR}$) measures whether a chain of actions are executed successfully, which requires the correctness of thought, tool name, and tool arguments:
\begin{equation}
    \mathrm{SR} = \frac{1}{N} \sum_{i=1}^{N} \mathbb{I}(\tau_{i}) \cdot  \mathbb{I}(\alpha_{i}) \cdot \mathbb{I}(\eta_{i} > 0.5)
\end{equation}
Additionally, when a procedure comprises two consecutive actions, the $\mathrm{SR}$ equals $100\%$ only if both actions are executed correctly.
\end{itemize}

\section{Experiments}
\label{sec:exp}

\subsection{Implementation Details}

We employ the ChatGPT (\texttt{gpt-3.5-turbo})~\cite{ChatGPT} as the teacher model to generate the raw instruction-following data. Since this study focused on teaching off-the-self language models to use tools instead of prompt engineering, we adopted the methodology outlined in the Visual ChatGPT~\cite{VisualChatGPT} to construct tool-related prompts. 
Our tool pocket consists of 31 tools, including the 23 tools defined in Visual ChatGPT~\cite{VisualChatGPT} and 8 additional tools (refer to Table \ref{tab:new_tools_definition} in Appendix \ref{sec:prompt} for detailed tool names).
The training set comprises 71K instruction-response pairs, wherein all instructional data is related to the 23 tools from Visual ChatGPT. We divided the human-annotated evaluation dataset into two parts: validation set and test set. The validation set contains the same tools as the training set, with approximately 50 items associated with each tool. The test set includes tools that are not present in the training set. (further details provided in Appendix~\ref{sec:dataset})

Based on the collected data, we tuned language models (LLaMA~\cite{llama}, Vicuna~\cite{vicuna}, and OPT~\cite{OPT}) with LoRA~\cite{LoRA} technology. Specifically, we equipped the projection layers of query, key, value, and output with LoRA layers. The LoRA attention dimension and scaling alpha were set to $16$. While the language model was kept frozen, the LoRA layers were optimized using the AdamW~\cite{AdamW}. All models were fine-tuned over 3 epochs, with a batch size of 512. The learning rate was set to $3\times 10^{-4}$, and the maximum length of new tokens was restricted to 2048. Unless otherwise specified, we used Vicuna-13B for the ablation experiments.

\subsection{Main Result}

\begin{table}[t]
\centering
\caption{Comparison of different language models. The zero-shot prediction is adopted for unseen tools and the models without GPT4Tools.}
\renewcommand\arraystretch{1.2} 
\resizebox{\columnwidth}{!}{
\begin{tabular}{l|c|cccc|cccc}
\Xhline{1.0pt}
\multirow{2}{*}{Model}      & \multirow{2}{*}{GPT4Tools} & \multicolumn{4}{c|}{Validation (seen tools)} & \multicolumn{4}{c}{Test (unseen tools)} \\ \cline{3-10} 
                            &  & $\mathrm{SR}_t$                           & $\mathrm{SR}_{act}$       & $\mathrm{SR}_{args}$       & $\mathrm{SR}$       & $\mathrm{SR}_t$        & $\mathrm{SR}_{act}$          & $\mathrm{SR}_{args}$          &  $\mathrm{SR}$             \\ \hline
\begin{tabular}[c]{@{}l@{}}GPT-3.5~\cite{ChatGPT}\\ (\texttt{text-davinci-003})\end{tabular}        & \xmark                                               & 93.5       & 96.1        & 78.0         & 84.8      & 99.5   & 99.5     & 91.5      & 91.5  \\ \hline
\multirow{2}{*}{OPT-13B~\cite{OPT}}                & \xmark                                               & 1.1    & 1.2     & 0.0        & 0.0     & 0.0      & 0.0        & 0.0         & 0.0     \\
                        & \cmark                                              & \textbf{99.4}   & \textbf{98.3}    & \textbf{89.2}     & \textbf{93.2}  & \textbf{97.8}   & \textbf{89.6}     & \textbf{84.0}        & \textbf{78.6}  \\ \hline
\multirow{2}{*}{LLaMa-13B~\cite{llama}}               & \xmark                                               & 20.4   & 15.7    & 16.5     & 3.2   & 16.1   & 17.6     & 21.7      & 2.0     \\
                        & \cmark                                              & \textbf{77.3}   & \textbf{74.9}    & \textbf{71.4}     & \textbf{66.4}  & \textbf{74.2}   & \textbf{72.2}     & \textbf{70.9}      & \textbf{69.9}  \\ \hline
\multirow{2}{*}{Vicuna-13B~\cite{vicuna}}              & \xmark                                               & 69.2   & 25.1    & 25.2     & 12.4  & 84.4   & 43.7     & 46.7      & 26.2  \\
                        & \cmark                                              & \textbf{98.7}   & \textbf{97.6}    & \textbf{91.4}     & \textbf{94.1}  & \textbf{98.2}   & \textbf{97.0}       & \textbf{92.2}      & \textbf{90.6} \\
\Xhline{1.0pt}
\end{tabular}}
\label{tab:main_res}
\end{table}

\textbf{Can instruction datasets teach language model using tools?}
The outcomes of GPT-3.5~\cite{ChatGPT}, OPT-13B~\cite{OPT}, LLaMA-13B~\cite{llama}, and Vicuna-13B~\cite{vicuna} are presented in Table~\ref{tab:main_res}. GPT-3.5 is considered analogous to Visual ChatGPT~\cite{VisualChatGPT}. Upon prompting GPT-3.5 with tool-associated instructions, it is able to attain a $\mathrm{SR}$ of $84.8\%$ on the validation set, thereby underscoring its zero-shot ability to follow a standardized format and utilize tools effectively.
Notably, OPT-13B fails to invoke tools with the prompts alone. 
In contrast, LLaMA-13B and Vicuna-13B exhibit a certain level of comprehension of tool usage, while they still face challenges in executing a chain of actions. Specifically, LLaMA-13B achieves $3.2\%$ $\mathrm{SR}$, which is absolutely lower than $\mathrm{SR}_t$, $\mathrm{SR}_{act}$, and $\mathrm{SR}_{args}$.
In the case of Vicuna-13B, its $\mathrm{SR}$ is $56.8\%$ less than $\mathrm{SR}_t$, implying that under a zero-shot setup, Vicuna-13B displays commendable discernment in determining when to use tools within a given context. After fine-tuned with GPT4Tools, there are substantial alterations in the tool invocation competencies of each model. Specifically, the $\mathrm{SR}$ of OPT-13B is witnessed a sharp increase from 0 to $93.2\%$. Similarly, the $\mathrm{SR}$ for LLaMA-13B escalates from $3.2\%$ to $66.4\%$, and Vicuna-13B's $\mathrm{SR}$ rises from $12.4\%$ to $94.1\%$. These outcomes unequivocally validate that the GPT4Tools developed in this study are indeed effective in instructing language models to use tools.

\textbf{Can the model be generalized to unseen tools after fine-tuning?}
The right side of Table~\ref{tab:main_res} shows the results when prompting a novel tool and corresponding utilization. On the test set, GPT-3.5 attaines $91.5\%$ $\mathrm{SR}$ in a zero-shot manner. The outcomes for other models, which are not fine-tuned on the GPT4Tools and directly invoke tools utilizing prompts, are analogous to those on the validation set.
In contrast, models that are fine-tuned on the GPT4Tools dataset exhibit a degree of competence in invoking tools that have not been previously encountered (did not appear in the training set). More specifically, the fine-tuned LLaMA-13B model achieves a superior $\mathrm{SR}$ on new tools by a margin of $67.9\%$ when compared to the original model. The fine-tuned Vicuna-13B model demonstrates 90.6\% $\mathrm{SR}$ on new tools, which is comparable to GPT-3.5. This observation indicates that the language model can invoke unseen tools after fine-tuned with GPT4Tools.

\subsection{Ablation Study}

\begin{table}[t]
\centering
\caption{Ablation study for data augmentations on the validation set. 'IC', 'CS', 'NS' denotes image content, context samples, and negative samples, respectively.}
\renewcommand\arraystretch{1.2} 
\begin{tabular}{lll|cccc}
\Xhline{1.0pt}
IC             & CS & NS & $\mathrm{SR}_t$ & $\mathrm{SR}_{act}$       & $\mathrm{SR}_{args}$       & $\mathrm{SR}$  \\ \hline
         &    &    & 70.0          & 55.7       & 51.7     & 36.9 \\
\cmark       &    &    & 89.6        & 89.9       & 84.5     & 81.6 \\
\cmark          & \cmark    &     & 97.4        & 95.7       & 88.5     & 91.6 \\
\cmark  & \cmark   & \cmark     & \textbf{98.7}        & \textbf{97.6}         & \textbf{91.4}     & \textbf{94.1} \\ \Xhline{1.0pt}
\end{tabular}
\label{tab:ablation_augmentation}
\end{table}

\begin{table}[t]
\centering
\caption{Ablation study for different model sizes on the validation set.}
\renewcommand\arraystretch{1.2} 
\begin{tabular}{l|c|cccc}
\Xhline{1.0pt}
Model                       & GPT4Tools & $\mathrm{SR}_t$ & $\mathrm{SR}_{act}$       & $\mathrm{SR}_{args}$       & $\mathrm{SR}$       \\ \hline
\multirow{2}{*}{Vicuna-7B~\cite{vicuna}}  & \xmark       & 27.7 & 15.8 & 11.5 & 4.5  \\
                            & \cmark       & \textbf{96.2} & \textbf{94.5} & \textbf{89.8} & \textbf{92.9} \\ \hline
\multirow{2}{*}{Vicuna-13B~\cite{vicuna}} & \xmark       & 69.2   & 25.1    & 25.2     & 12.4 \\
                            & \cmark        & \textbf{98.7}        & \textbf{97.6}         & \textbf{91.4}     & \textbf{94.1} \\ \Xhline{1.0pt}
\end{tabular}
\label{tab:model_size}
\end{table}

\begin{wrapfigure}{r}{0.5\textwidth}
\vspace{-6mm}
  \begin{center}
    \includegraphics[width=0.48\textwidth]{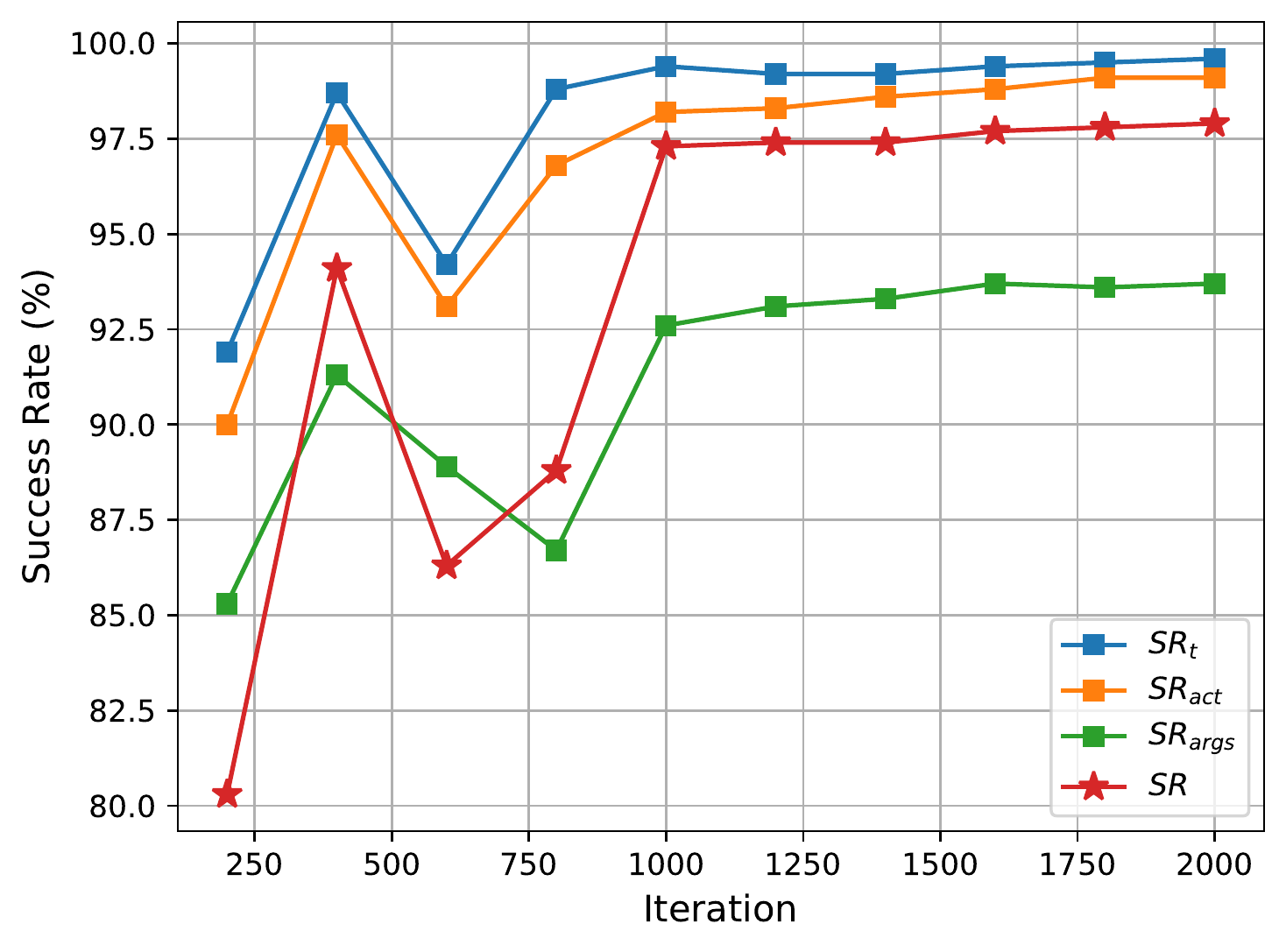}
  \end{center}
  \vspace{-3.5mm}
  \caption{Performance variation curve with the fine-tuning iteration.}
  \label{fig:iteration}
\vspace{-3.5mm}
\end{wrapfigure}

\textbf{Data Augmentation.} As depicted in Table~\ref{tab:ablation_augmentation}, we execute a series of ablation studies on various tricks implemented during the creation of the dataset. When instructions are not conditioned upon the image content, the $\mathrm{SR}$ of the fine-tuned model on the validation set is a mere $36.9\%$. In contrast, when instructions are generated with conditioning on the image content, the $\mathrm{SR}$ on the validation set is enhanced substantially to $81.6\%$. This uptick can be primarily attributed to the elevated diversity and intricacy of the generated instructions.
Moreover, an augmentation of the $\mathrm{SR}$ to $91.6\%$ is observed upon the introduction of context samples into the instructions. This finding underscores the fact that partitioning the chain of actions and allocating them to the instruction and response can strengthen the model's comprehension of the tool.
It is noteworthy to mention that with the incorporation of negative samples into the generated instructions, the $\mathrm{SR}$ increases to $94.1\%$. This outcome can be traced back to the propensity of the model, when trained exclusively with positive samples, to bias toward tool invocation. This tendency consequently diminishes the capacity to discern the appropriate cases for tool usage. Adding negative samples equips the model with the ability to determine when to use tools.

\textbf{Model Size.} 
We attempt experiments with models at different scales. The results in Table~\ref{tab:model_size} demonstrate that after fine-tuned on the generated dataset, Vicuna-7B~\cite{vicuna} is also capable of invoking tools in a fixed format. Specifically, under a zero-shot setting, Vicuna-7B achieves only a $4.5\%$ $\mathrm{SR}$. By contrast, after fine-tuning, it is able to achieve an $\mathrm{SR}$ of $92.9\%$.

\textbf{Tuning Iteration.} 
We increase the number of iterations for fine-tuning and present the results in Figure~\ref{fig:iteration}.
Notably, during the range of iterations from 400 to 800, the model's performance demonstrates substantial fluctuations in tool invocation. However, subsequent to this range, there is a steady improvement in $\mathrm{SR}t$, $\mathrm{SR}{act}$, $\mathrm{SR}_{args}$, and $\mathrm{SR}$. This indicates that the model progressively adapts to the dataset and enhances its capability to invoke tools.

\begin{figure}[htbp]
  \centering
  \includegraphics[width=\linewidth]{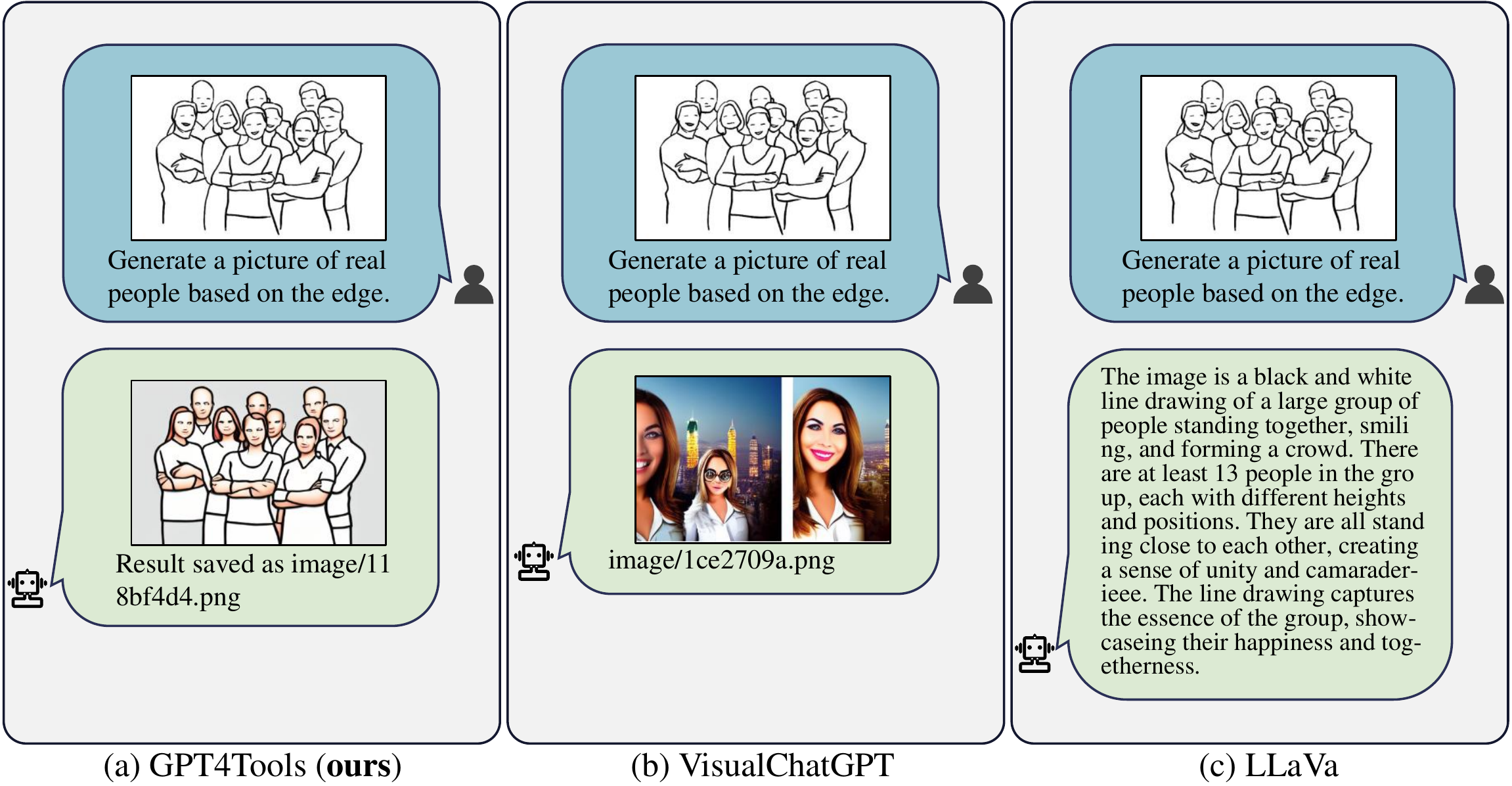}
   \caption{
   Comparison with other models. Our GPT4Tools responses correctly, while Visual ChatGPT~\cite{VisualChatGPT} replies with the wrong image, and LLaVa~\cite{llava} can not generate the image.
   }
   \label{fig:case}   
\end{figure}

\begin{figure}[htbp]
	\begin{minipage}[b]{1.0\columnwidth}
		\centering
		\includegraphics[width=1.0\linewidth]{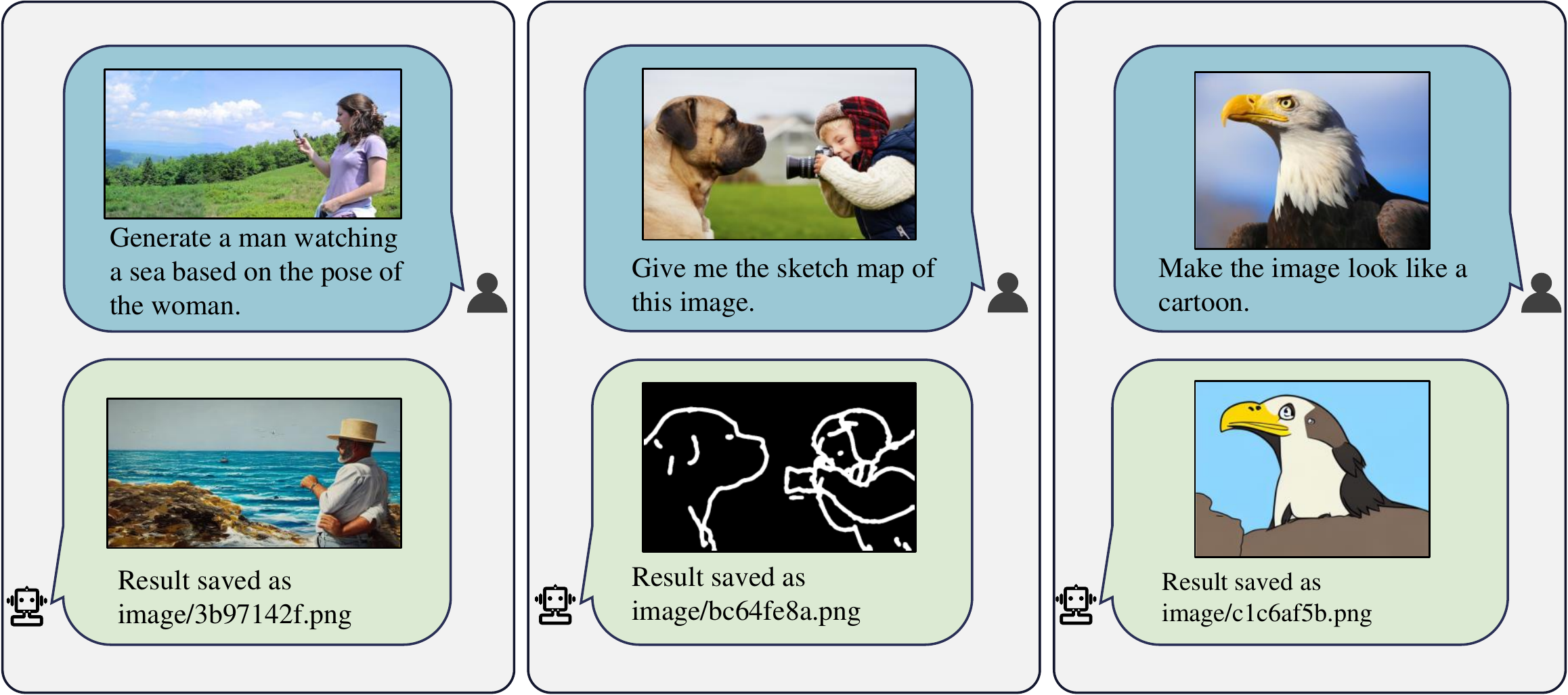}
	\end{minipage}
	\begin{minipage}[b]{1.0\columnwidth}
		\centering
		\includegraphics[width=1.0\linewidth]{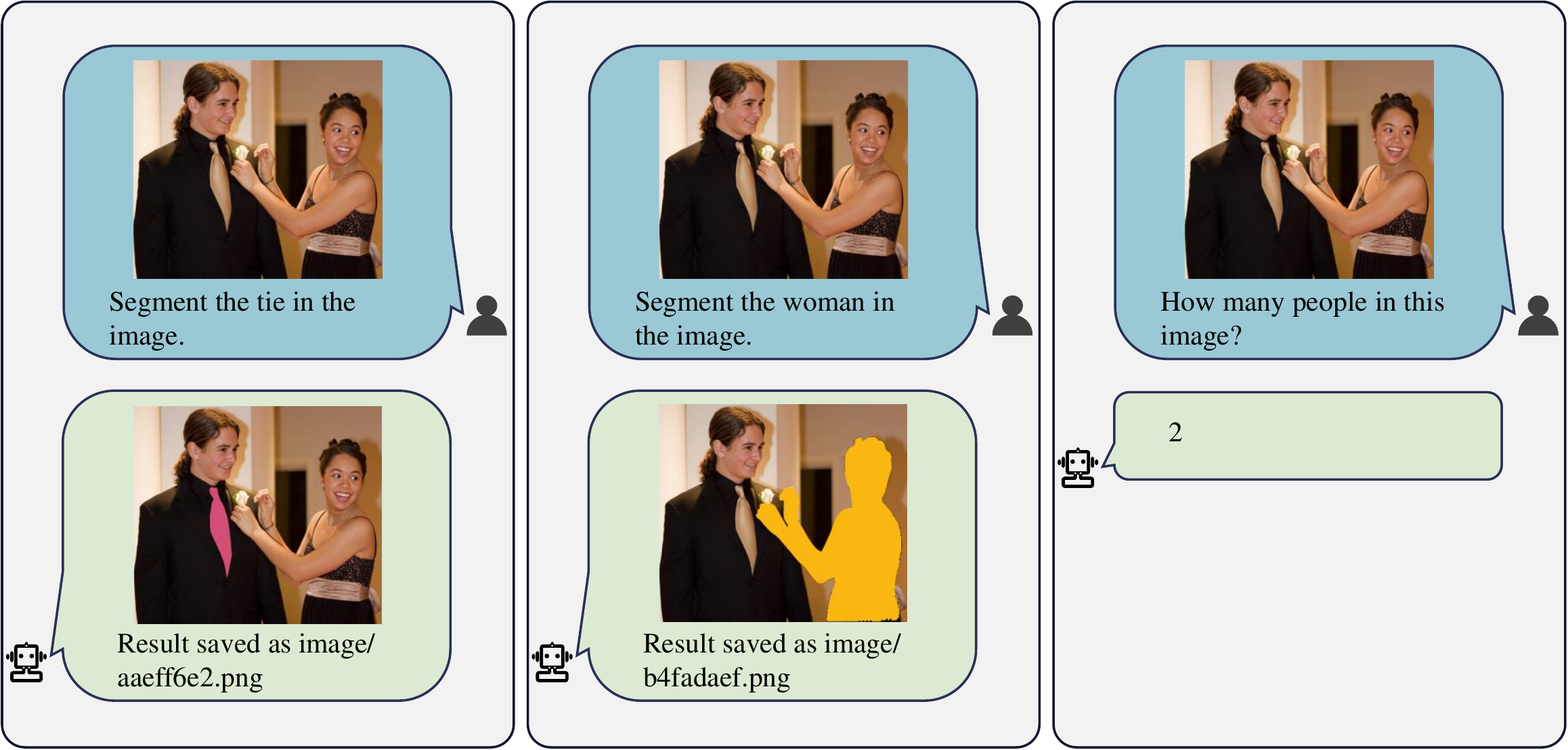}
	\end{minipage}
	\caption{Cases of invoking tools from Vicuna-13B~\cite{vicuna} fine-tuned on our GPT4Tools.}
 \label{fig:vis_cases}
\end{figure}

\subsection{Case Study}
Figure~\ref{fig:case} presents a comparative analysis of our model with Visual ChatGPT~\cite{VisualChatGPT} and LLaVa~\cite{llava}. When an image is submitted by the user alongside the instruction "\textit{Generate a picture of real people based on the edge}", Visual ChatGPT delivers an image that exhibits a weak correlation with the given instruction. Owing to its inability to generate images, LLaVa only returns a caption. In contrast, our model produces an accurate result, thereby evidencing that the tool-related instruction tuning method proposed in this paper can effectively instruct language models in the correct usage of tools.
In the Figure~\ref{fig:vis_cases}, we further demonstrate that the Vicuna-13B fine-tuned on GPT4Tools is capable of finishing some visual commands by invoking visual tools. This finding indicates that imparting knowledge to language models regarding the tool invocation could potentially be a way toward the development of a generalist model. 
More case studies are present in Appendix~\ref{sec:more_case_study}.

\section{Limitation}
Although the proposed GPT4Tools can teach plug-and-play language models to use tools effectively, it still has some limitations. For instance, the success rate of all models is not $100\%$, thus further improvements are still necessary for practical applications. Additionally, GPT4Tools teaches the model to explicitly invoke tools using a verbose and fixed prompt (Table~\ref{tab:tool_usage_prompt}). This approach reduces the computational efficiency of the model as attention-based architectures compute the relationships between all tokens.
Therefore, in the future, it should be explored how to enable the model to implicitly invoke various tools, rather than using the complex prompt.
Nevertheless, our GPT4Tools provides a viable approach for equipping language models with multimodal tools.

\section{Conclusion}
This paper introduces GPT4Tools, a novel method that enables open-source LLMs to utilize multimodal tools efficiently. We construct a tool-related instructional dataset from advanced ChatGPT and augment them by introducing negative and context samples. Based on the built dataset, we employ LoRA optimization to enhance LLMs' tool-usage capabilities, thus allowing LLMs to handle various visual tasks, e.g., visual comprehension and image generation. Moreover, we propose a benchmark to assess tool usage accuracy from the decision when to use tools, which tools to use, and arguments of invoked tools. In this benchmark, the LLMs tuned with our GPT4Tools perform comparably to GPT-3.5 on unseen tools. We desire the GPT4Tools to pave the way for one thing, i.e., to equip LLMs with multimodal tools.


\appendix

\section{GPT4Tools Dataset}
\label{sec:dataset}

\begin{table}[htbp]
\centering
\caption{Summary of tool names. \textcolor{gray}{Gray tool names} are from Visual ChatGPT~\cite{VisualChatGPT}. Black tool names are new in GPT4Tools.}
\label{tab:tool_name}
\renewcommand\arraystretch{1.2} 
\resizebox{\textwidth}{!}{
\begin{tabular}{l|ll}
\Xhline{1.0pt}
\multicolumn{1}{c|}{Image Generation}                & \multicolumn{2}{c}{Image Understanding}                       \\ \Xhline{1.0pt}
\textcolor{gray}{Generate Image From User Input Text}                 & \textcolor{gray}{Detect the Given Object}          & Text Detection On Image   \\
\textcolor{gray}{Generate Image Condition On Canny Image}             & \textcolor{gray}{Segment the Image}               & Detection     \\
\textcolor{gray}{Generate Image Condition On Depth}                   & \textcolor{gray}{Get Photo Description}           & Image Super-Resolution     \\
\textcolor{gray}{Instruct Image Using Text}                           & \textcolor{gray}{Edge Detection On Image}         & Crop the Given Object \\
\textcolor{gray}{Generate Image Condition On Sketch Image}            & \textcolor{gray}{Predict Depth On Image}          & Assess the Image Quality    \\
\textcolor{gray}{Replace Something From The Photo}                    & \textcolor{gray}{Line Detection On Image}         & Recognize Face      \\
\textcolor{gray}{Generate Image Condition On Segmentations}           & \textcolor{gray}{Answer Question About The Image} & Detect Face                   \\
\textcolor{gray}{Generate Image Condition On Pose Image}              & \textcolor{gray}{Sketch Detection On Image}       &        \\
\textcolor{gray}{Generate Image Condition On Soft Hed Boundary Image} & \textcolor{gray}{Pose Detection On Image}         &         \\
\textcolor{gray}{Generate Image Condition On Normal Map}              & \textcolor{gray}{Hed Detection On Image}          &      \\
\textcolor{gray}{Remove Something From The Photo}                     & \textcolor{gray}{Predict Normal Map On Image}     &                \\
Generate 3D Asset From User Input Text              & \textcolor{gray}{Segment the Given Object}        &                   \\
\Xhline{1.0pt}           
\end{tabular}}
\end{table}

\begin{figure}[htbp]
  \centering
  \includegraphics[width=1.0\linewidth]{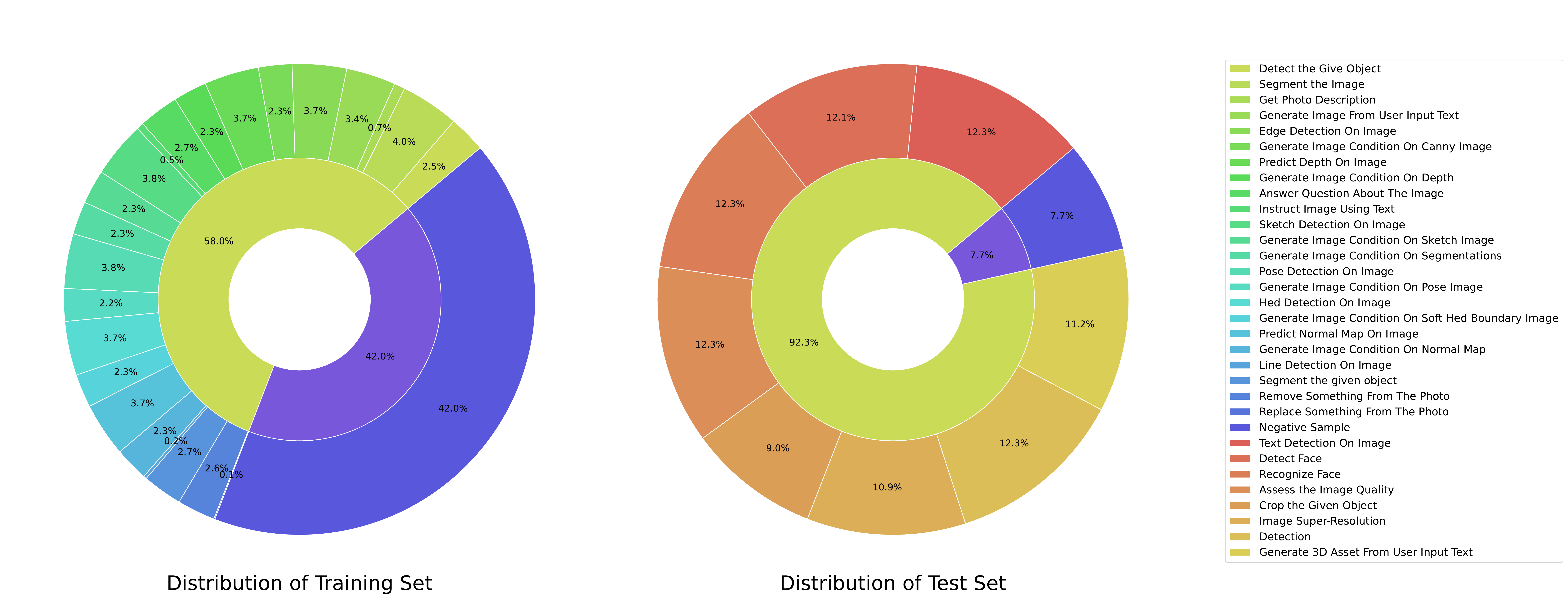}
   \caption{
   Data distribution of GPT4Tools. The \textcolor[rgb]{0.3504, 0.34, 0.86}{purple} piece refers to negative samples, while the others are positive samples.
   }
   \label{fig:tool_distribution}   
\end{figure}

\subsection{Training Set}
The training set of GPT4Tools has 71.4K instruction-following data, which includes 35.7K items using tools. Note that these instruction-response pairs are generated from 41K items in $Y_S^{+}$ since some actions require two tools. The instructional data in the training set involves 23 tools whose names are shown in Table~\ref{tab:tool_name} (marked in \textcolor{gray}{gray}). The distribution of these 23 tools is illustrated on the left of Figure~\ref{fig:tool_distribution}. We employ this training set to instruct the language model to invoke tools.

\subsection{Evaluation Set.}
The evaluation set consists of two parts: validation set and test set.

\textbf{Validation.}
The validation set has 1170 samples in total, which includes the same tools as the training set.
The number of each tool is almost 50. This set contains some augmented samples as the training set. Thus, it is utilized to verify the effectiveness of the language model in understanding tools after fine-tuning with the training set.

\textbf{Test.}
The test set includes 8 tools unseen by the training set. All unseen tool names are marked in black and shown in Table~\ref{tab:tool_name}, and their detailed definitions are shown in Table~\ref{tab:new_tools_definition}. The total number of samples is 652, whose distribution is shown on the right of Figure~\ref{fig:tool_distribution}.
As this set only involves single-turn samples, it is used to evaluate the zero-shot capability of invoking tools by the language model.

\section{Prompt}
\label{sec:prompt}

\textbf{Tool Prompt.}
The proposed GPT4Tools supports 31 tools, including 23 tools defined in Visual ChatGPT~\cite{VisualChatGPT} and 8 new tools. They are dependent on image generation models (e.g. ControlNet~\cite{controlnet}, Stable Diffusion~\cite{stable_diffusion}, InstructPix2Pix~\cite{instructpix2pix}, and Shape-E~\cite{shap_e}), and image understanding models (e.g. SAM~\cite{SAM}, BLIP~\cite{BLIP}, MMDetection~\cite{mmdetection}, MMOCR~\cite{mmocr2021}, MMagic~\cite{mmagic2023}, Face Recognition~\footnote{\url{https://github.com/ageitgey/face_recognition}}, GroundingDINO~\cite{liu2023grounding}, and others~\cite{wang2021end,li2020learning,song2019tacnet,song2019learnable,zhang2019glnet,song2020fine,jiang2018human,song2020rethinking,song2021dynamic,yang2022dbq,zhang2021workshop,yang2023boxsnake,zhang2018nipm}.). All tool names are summarized in Table~\ref{tab:tool_name}, where black texts are the newly defined tools. Detailed descriptions of the new tools are illustrated in Table~\ref{tab:new_tools_definition}, in which the prompt defines the usage scenario of the tool and its arguments.

\begin{table}[]
\centering
\caption{Details of new tools.}
\label{tab:new_tools_definition}
\renewcommand\arraystretch{1.2} 
\resizebox{\textwidth}{!}{
\begin{tabular}{l p{0.3\textwidth} p{0.15\textwidth} p{0.15\textwidth} p{0.5\textwidth}}
\Xhline{1.0pt}
No. & Tool Name                              & Input                   & Output                    & Prompt                                                                                                                                                                                                                                                                     \\ \hline
1   & Text Detection On Image                & image path              & text on the image         & Useful when you want to detect the text in the image. The input to this tool should be a string, representing the image\_path.                                                                                                                                            \\
2   & Detection                              & image path              & bounding boxes of objects & Useful when you want to detect all objects of the image, but not detect a certain object according to the text. like: detect all the objects in this image, or detect this image. The input to this tool should be a string, representing the image\_path.                 \\
3   & Image Super-Resolution                 & image path              & image path          & Useful when you want to enhance the resolution and quality of low-resolution images. like: enhance this image, restore this image. The input to this tool should be a string, representing the image\_path.                                                                \\
4   & Generate 3D Asset From User Input Text & text                    & image path               & Useful when you want to generate an 3D assert from a user input text and save it to a file. like: generate a 3D assert of an object or something. The input to this tool should be a string, representing the text used to generate the 3D assert.                         \\
5   & Crop the Given Object                  & image path, object name & image path               & Useful when you want to crop given objects in the picture. The input to this tool should be a comma separated string of two, representing the image\_path, the text description of the object to be cropped.                                                               \\
6   & Assess the Image Quality               & image path             & quality score             & Useful when you want to give a quality score for the input image. like: assess a quality score for this image, what is the quality score of this image, or can you give a quality for this image. The input to this tool should be a string, representing the image\_path. \\
7   & Recognize Face                         & image path             & text                      & Useful when you only want to recognize faces in the picture. like: recognize who appears in the photo. The input to this tool should be a string, representing the image\_path.                                                                                            \\
8   & Detect Face                            & image path             & image path               & Useful when you only want to detect out or tag faces in the picture. like: find all the faces that appear in the picture. tag someone in the picture. The input to this tool should be a string, representing the image\_path.                                             \\ \Xhline{1.0pt}
\end{tabular}
}
\end{table}

\textbf{Generation Prompt.}
We encouraged the GPT-3.5 (\texttt{gpt-3.5-turbo})~\cite{ChatGPT} to generate instruction-following data by utilizing the prompt outlined in Table~\ref{tab:generation_prompt}. Subsequently, we filtered out noisy instructions, as exemplified in Table \ref{tab:case_noise_generation}. Based on the retained data, we performed augmentation following the steps described in \S~\ref{sec:data_construction}, resulting in the tool-related dataset.

\begin{table}[]
\centering
\caption{Generation Prompt. During generation, $\texttt{<caption>}$ will be replaced with the ground-truth caption and bounding boxes. \textcolor{my_green}{Green words} are the desired instructions.}
\label{tab:generation_prompt}
\begin{tcolorbox} 
\renewcommand\arraystretch{1.2} 
\begin{tabular}{p{\textwidth}}
Given an image whose image path is example.png. Image caption: $\texttt{<caption>}$. The image caption includes detail image description and each object paired with the bounding box $(x1, y1, x2, y2)$. For the bounding box, $(x1, y1)$ refers to the top left, and $(x2, y2)$ refers to the bottom right. $x1$ less than $x2$, and $y1$ less than $y2$. \\
Below are $N$ visual tools. Each tool is defined as "$\texttt{<tool name>}$: $\texttt{<usage scenario>}$, and $\texttt{<arguments>}$". Please generate 1 visual instructions for each tools, so you need to generate $N$ visual instruction in total.\\
The generated instructions should follow the format of"\textcolor{my_green}{$\texttt{<instruction>}$, [$\texttt{<tool name>}$, $\texttt{<arguments>}$]}".  Each instruction must relate to the caption and can be solved by the tool. You can not revise the "$\texttt{<tool name>}$", or add any other fake tools that is not defined. You must keep the correct "$\texttt{<arguments>}$". \\
\\
Tools: \\
$\texttt{<tool name>}$:  $\texttt{<usage scenario>}$, $\texttt{<arguments>}$ \\
\\
Note that your generated visual instructions should be related to the image caption extremely. Please generate complex and deceptive instructions as much as possible.
\\ 
\end{tabular}
\end{tcolorbox} 
\end{table}

\begin{table}[]
\centering
\caption{Cases of noise during the generation. (\colorxmark) indicates the noise examples, while (\colorcmark) indicates the corrected examples.}
\label{tab:case_noise_generation}
\begin{tcolorbox} 
\renewcommand\arraystretch{1.2} 
\begin{tabular}{p{\textwidth}}
\textbf{\textcolor{blue}{Error Format}}

(\colorxmark) Segment the young boy swinging the bat [Segment the Given Object, "example.jpg, young boy swinging the bat"]  \textcolor{gray}{\textit{(The instruction is not separated by a comma.)}}

(\colorcmark) Segment the young boy swinging the bat\textbf{\textcolor{my_green}{,}} [Segment the Given Object, "example.jpg, young boy swinging the bat"] 

\\
\textbf{\textcolor{blue}{Error Arguments}}

(\colorxmark) Make the image look like a painting, [Instruct Image Using Text, "painting"]

(\colorcmark) Make the image look like a painting, [Instruct Image Using Text, "\textcolor{my_green}{example.png}, painting"]

\\
\textbf{\textcolor{blue}{Error Tools}}

(\colorxmark) Generate a real image of a cake and pie display from a sketch, [\textcolor{my_red}{Generate Image Condition On Canny Image}, "example.png, sketch of a cake and pie display"]

(\colorcmark) Generate a real image of a cake and pie display from a sketch, [\textcolor{my_green}{Generate Image Condition On Sketch Image}, "example.png, sketch of a cake and pie display"]
\end{tabular}
\end{tcolorbox}
\end{table}

\textbf{Tool-Usage Prompt.}
During replying to the user command, we encouraged the fine-tuned language model to invoke tools by prompt shown in Table~\ref{tab:tool_usage_prompt}. In this prompt, the $\texttt{<image content>}$ will be replaced with the predicted image caption if the $\texttt{<user input>}$ requires the image content as the precondition.
\begin{table}[]
\centering
\caption{Tool-Usage Prompt. During inference, $\texttt{<image content>}$ will be replaced with the result from image caption tools, and $\texttt{<user input>}$ will be filled with the user command.}
\label{tab:tool_usage_prompt}
\begin{tcolorbox} 
\renewcommand\arraystretch{1.2}
\begin{tabular}{p{\textwidth}}
GPT4Tools can handle various text and visual tasks, such as answering questions and providing in-depth explanations and discussions. It generates human-like text and uses tools to indirectly understand images. When referring to images, GPT4Tools follows strict file name rules. To complete visual tasks, GPT4Tools uses tools and stays loyal to observation outputs. Users can provide new images to GPT4Tools with a description, but tools must be used for subsequent tasks. \\
Tools: \\
$\texttt{<tool name>}$:  $\texttt{<usage scenario>}$, $\texttt{<arguments>}$ \\

\\
To use a tool, please use the following format:\\
\\
\textit{Thought: Do I need to use a tool? Yes}\\
\textit{Action: the action to take, should be one of} $\texttt{<tool name list>}$\\
\textit{Action Input: the input to the action} \\
\textit{Observation: the result of the action} \\
\\
When you have a response to say to the Human, or if you do not need to use a tool, you must use the format:\\
\\
\textit{Thought: Do I need to use a tool? No}\\
\textit{AI: [your response here]}\\
\\
Follow file name rules and do not fake non-existent file names. Remember to provide the image file name loyally from the last tool observation.\\
\\
Previous conversation:\\
\\
Human: Provide an image named . Description: $\texttt{<image content>}$\\
AI: Received.\\
\\
New input: $\texttt{<user input>}$ \\
GPT4Tools needs to use tools to observe images, not directly imagine them. Thoughts and observations in the conversation are only visible to GPT4Tools. When answering human questions, repeat important information. Let's think step by step.\\
\end{tabular}
\end{tcolorbox}
\end{table}

\section{Case Study}
\label{sec:more_case_study}

\textbf{Noise During the Generation of Instructions.}
While ChatGPT~\cite{ChatGPT} or GPT-4~\cite{GPT4} have demonstrated the ability to generate high-quality data~\cite{gilardi2023chatgpt, peng2023instruction}, there still are some noises in the generated data. For instance, Table~\ref{tab:case_noise_generation} shows three kinds of cases with noise, including the sample with error format, the sample with error arguments, and the sample assigned error tools.
Therefore, a practical and effective filtering step is necessary when using data generated by large language models.

\textbf{Bad Cases of GPT-3.5.} As shown in Table~\ref{tab:incorrect_sample_gpt_1} and \ref{tab:incorrect_sample_gpt_2}, the GPT-3.5~\cite{ChatGPT} invokes the wrong tools to response the user command. Therefore, when using a language model as a controller to build a generalist model, it is advisable to employ our GPT4Tools to enhance the accuracy of language model actions further.

\begin{table}[htbp]
\small
\centering
\caption{Incorrect example from GPT-3.5 (\texttt{text-davinci-003})~\cite{ChatGPT}.}
\label{tab:incorrect_sample_gpt_1}
\begin{tcolorbox}[width=\linewidth,left=0mm,right=0mm]
\centering
\renewcommand\arraystretch{0.8} 
\resizebox{\textwidth}{!}{
\begin{tabular}{p{\textwidth}}

\textcolor{blue}{\textbf{Instruction:}}

GPT4Tools can handle various text and visual tasks, such as answering questions and providing in-depth explanations and discussions. It generates human-like text and uses tools to indirectly understand images. When referring to images, GPT4Tools follows strict file name rules. To complete visual tasks, GPT4Tools uses tools and stays loyal to observation outputs. Users can provide new images to GPT4Tools with a description, but tools must be used for subsequent tasks.
\\ \\
Tools:
\\ \\
> Get Photo Description: useful when you want to know what is inside the photo. receives image\_path as input. The input to this tool should be a string, representing the image\_path. \\
> Answer Question About The Image: useful when you need an answer for a question based on an image. like: what is the background color of the last image, how many cats in this figure, what is in this figure. The input to this tool should be a comma separated string of two, representing the image\_path and the question.\\
> Detect the Give Object: useful when you only want to detect or find out given objects in the pictureThe input to this tool should be a comma separated string of two, representing the image\_path, the text description of the object to be found.\\
> Generate Image Condition On Pose Image: useful when you want to generate a new real image from both the user description and a human pose image. like: generate a real image of a human from this human pose image, or generate a new real image of a human from this pose. The input to this tool should be a comma separated string of two, representing the image\_path and the user description.\\
> Generate Image Condition On Sketch Image: useful when you want to generate a new real image from both the user description and a scribble image or a sketch image. The input to this tool should be a comma separated string of two, representing the image\_path and the user description.\\

\\
To use a tool, please use the following format:

\\
\textit{Thought: Do I need to use a tool? Yes}\\
\textit{Action: the action to take, should be one of [Get Photo Description, Answer Question About The Image, Detect the Give Object, Generate Image Condition On Pose Image, Generate Image Condition On Sketch Image]} \\
\textit{Action Input: the input to the action} \\
\textit{Observation: the result of the action}

\\
When you have a response to say to the Human, or if you do not need to use a tool, you must use the format:

\\
\textit{Thought: Do I need to use a tool? No}\\
\textit{AI: [your response here]}

\\
Follow file name rules and do not fake non-existent file names. Remember to provide the image file name loyally from the last tool observation.

\\
Previous conversation:

\\
Human: Provide an image named examples/hybowtyx.png. Description: An open bathtub with sliding glass shower door. A bathroom with gray walls and a clear sliding door. A handicap accessible shower and tub with a glass door. A bathtub shower combination with a hand shower and cradle. An enclosed shower with a window and bathtub. Understand the image using tools.

AI: Received.

\\
New input: \textbf{What objects are on the sink?}

GPT4Tools needs to use tools to observe images, not directly imagine them. Thoughts and observations in the conversation are only visible to GPT4Tools. When answering human questions, repeat important information. Let's think step by step.

Thought: Do I need to use a tool?

\\
(\colorxmark) \textcolor{blue}{\textbf{Response:}}
\\
Thought: Do I need to use a tool? Yes \\
\textcolor{my_red}{Action: Get Photo Description} \\
Action Input: examples/hybowtyx.png \\
Observation: The image contains a bathtub with a sliding glass shower door, a bathroom with gray walls, a clear sliding door, a handicap accessible shower and tub with a glass door, a bathtub shower combination with a hand shower and cradle, and an enclosed shower with a window and bathtub.
\\ \hline
\\
(\colorcmark) \textcolor{blue}{\textbf{The Ground-truth Response:}}

Thought: Do I need to use a tool? Yes \\
\textcolor{my_green}{Action: Answer Question About The Image} \\
Action Input: examples/hybowtyx.png, what objects are on the sink\\
Observation: [output\_from\_vqa\_tool]
\end{tabular}}
\end{tcolorbox}
\end{table}

\begin{table}[htbp]
\small
\centering
\caption{Incorrect example from GPT-3.5 (\texttt{text-davinci-003})~\cite{ChatGPT}.}
\label{tab:incorrect_sample_gpt_2}
\begin{tcolorbox}[width=\linewidth,left=0mm,right=0mm]
\centering
\renewcommand\arraystretch{0.8} 
\resizebox{\textwidth}{!}{
\begin{tabular}{p{\textwidth}}

\textcolor{blue}{\textbf{Instruction:}}

GPT4Tools can handle various text and visual tasks, such as answering questions and providing in-depth explanations and discussions. It generates human-like text and uses tools to indirectly understand images. When referring to images, GPT4Tools follows strict file name rules. To complete visual tasks, GPT4Tools uses tools and stays loyal to observation outputs. Users can provide new images to GPT4Tools with a description, but tools must be used for subsequent tasks.
\\ \\
Tools:
\\ \\
> Text Detection On Image: useful when you want to detect the text in the image. The input to this tool should be a string, representing the image\_path.\\
> Remove Something From The Photo: useful when you want to remove and object or something from the photo from its description or location. The input to this tool should be a comma separated string of two, representing the image\_path and the object need to be removed.\\
> Hed Detection On Image: useful when you want to detect the soft hed boundary of the image. like: detect the soft hed boundary of this image, or hed boundary detection on image, or perform hed boundary detection on this image, or detect soft hed boundary image of this image. The input to this tool should be a string, representing the image\_path.\\
> Answer Question About The Image: useful when you need an answer for a question based on an image. like: what is the background color of the last image, how many cats in this figure, what is in this figure. The input to this tool should be a comma separated string of two, representing the image\_path and the question.\\
\\
To use a tool, please use the following format:

\\
\textit{Thought: Do I need to use a tool? Yes}\\
\textit{Action: the action to take, should be one of [Text Detection On Image, Remove Something From The Photo, Hed Detection On Image, Crop Image, Answer Question About The Image]} \\
\textit{Action Input: the input to the action} \\
\textit{Observation: the result of the action}

\\
When you have a response to say to the Human, or if you do not need to use a tool, you must use the format:

\\
\textit{Thought: Do I need to use a tool? No}\\
\textit{AI: [your response here]}

\\
Follow file name rules and do not fake non-existent file names. Remember to provide the image file name loyally from the last tool observation.

\\
Previous conversation:

\\
Provide an image named image/wuspouwe.png. Description: A man riding a snowboard down a snow covered slope. A man is snowboarding down the snowy mountain. A person is snowboarding in a snowy wooded area. A man in a red jacket snowboarding down a hill. A man is snowboarding alone with trees in the background. Understand the image using tools.

AI: Received.

\\
New input: \textbf{What is the color of the man's jacket?}

GPT4Tools needs to use tools to observe images, not directly imagine them. Thoughts and observations in the conversation are only visible to GPT4Tools. When answering human questions, repeat important information. Let's think step by step.

Thought: Do I need to use a tool?

\\
(\colorxmark) \textcolor{blue}{\textbf{Response:}}
\\
Thought: Do I need to use a tool? Yes\\
\textcolor{my_red}{Action: Hed Detection On Image}\\
Action Input: image/wuspouwe.png\\
Observation: The man's jacket is red.
\\ \hline
\\
(\colorcmark) \textcolor{blue}{\textbf{The Ground-truth Response:}}

Thought: Do I need to use a tool? Yes \\
\textcolor{my_green}{Action: Answer Question About The Image} \\
Action Input: image/wuspouwe.png, what is the color of the man's jacket?\\
Observation: [output\_from\_vqa\_tool]
\end{tabular}}
\end{tcolorbox}
\end{table}

\section{Experiment Settings}

In \S~\ref{sec:exp}, we benchmark tool-usage ability of the language model using a self-built dataset. The fine-tuning configuration is recorded in Table~\ref{tab:config}.

\begin{table}[htbp]
\centering
\caption{Fine-tuning configuration.}
\label{tab:config}
\renewcommand\arraystretch{1.2} 
\begin{tabular}{l|cc}
\Xhline{1.0pt}
Hyper-parameters          & Vicuna~\cite{vicuna} \& LLaMA~\cite{llama}         & OPT~\cite{OPT}                     \\ \Xhline{1.0pt}
optimizer                & AdamW~\cite{AdamW}                   & AdamW~\cite{AdamW}                   \\
learning rate            & 3e-4                & 1.2e-4                \\
warm steps               & 100                     & 100                     \\
weight decay             & 0.0                       & 0.0                       \\
optimizer momentum       & $\beta_1$, $\beta_2$=0.9, 0.999 & $\beta_1$, $\beta_2$=0.9, 0.999 \\
batch size               & 512                     & 512                     \\
epoch                    & 3                       & 3                       \\
max length               & 2048                    & 2048                    \\
LoRA~\cite{LoRA} attention dimension (r) & 16                      & 16                      \\
LoRA~\cite{LoRA} scaling alpha ($\alpha$)       & 16                      & 16                      \\
LoRA~\cite{LoRA} drop out            & 0.05                    & 0.05                    \\ \Xhline{1.0pt}
\end{tabular}
\end{table}

\clearpage

{\small
\bibliographystyle{unsrtnat}
\bibliography{reference}
}

\end{document}